\documentclass{article}

\usepackage{microtype}
\usepackage{graphicx}
\usepackage{subcaption}
\usepackage{booktabs} 
\usepackage{enumitem}

\usepackage{hyperref}

 \usepackage[accepted]{icml2024}

\usepackage{amsmath}
\usepackage{amssymb}
\usepackage{mathtools}
\usepackage{amsthm}

\usepackage[capitalize,noabbrev]{cleveref}

\theoremstyle{plain}

\theoremstyle{definition}

\theoremstyle{remark}

\usepackage[textsize=tiny]{todonotes}

\icmltitlerunning{Amazing Things Come From Having Many Good Models}

\begin{document}

\twocolumn[
\icmltitle{Amazing Things Come From Having Many Good Models}



\icmlsetsymbol{equal}{*}

\begin{icmlauthorlist}
\icmlauthor{Cynthia Rudin}{duke,equal}
\icmlauthor{Chudi Zhong}{duke}
\icmlauthor{Lesia Semenova}{duke}
\icmlauthor{Margo Seltzer}{ubc}
\icmlauthor{Ronald Parr}{duke}
\icmlauthor{Jiachang Liu}{duke}
\icmlauthor{Srikar Katta}{duke}
\icmlauthor{Jon Donnelly}{duke}
\icmlauthor{Harry Chen}{duke}
\icmlauthor{Zachery Boner}{duke}
\end{icmlauthorlist}

\icmlaffiliation{duke}{Department of Computer Science, Duke University, Durham, North Carolina, USA}
\icmlaffiliation{ubc}{Department of Computer Science, University of British Columbia, Vancouver, Canada}

\icmlcorrespondingauthor{Cynthia Rudin}{cynthia@cs.duke.edu}

\icmlkeywords{Rashomon Effect, Interpretability, Sparsity, Interactivity, Fairness}

\vskip 0.3in
]



\printAffiliationsAndNotice{$^{*}$ Authors, except the first, are listed in reverse alphabetical order, since there are many good ways to list equally contributing authors.}  

\begin{abstract}
The \textit{Rashomon Effect}, coined by Leo Breiman, describes the phenomenon that there exist many equally good predictive models for the same dataset. This phenomenon happens for many real datasets and when it does, it sparks both magic and consternation, but mostly magic. In light of the Rashomon Effect, this perspective piece proposes reshaping the way we think about machine learning, particularly for tabular data problems in the nondeterministic (noisy) setting. We address how the Rashomon Effect impacts (1) the existence of simple-yet-accurate models, (2) flexibility to address user preferences, such as fairness and monotonicity, without losing performance, (3) uncertainty in predictions, fairness, and explanations, (4) reliable variable importance, (5) algorithm choice, specifically, providing advanced knowledge of which algorithms might be suitable for a given problem, and (6) public policy. We also discuss a theory of when the Rashomon Effect occurs and why. Our goal is to illustrate how the Rashomon Effect can have a massive impact on the use of machine learning for complex problems in society.
\end{abstract}

\section{Introduction}
Real-world datasets often admit many approximately-equally-good models. Leo Breiman called this phenomenon the \textit{Rashomon Effect}, naming it after a Japanese movie in which four different views of the same murder show no single truth, just many reasonable explanations, for what happened \citep{breiman2001statistical,RashMovie}. 
One might think of the Rashomon Effect as a nuisance that prevents us from getting at a single ``true'' understanding of the data due to uncertainty, but from another perspective, the Rashomon Effect unlocks a treasure trove of information about the relationship of real datasets to families of predictive models. Harnessing this knowledge has massive practical implications, providing answers to some of the most fundamental questions in machine learning, such as: Is there an accuracy-interpretability trade-off?  Which algorithm(s) are suitable for a given dataset? How can we easily (i.e., without solving a difficult optimization problem) find a model that incorporates multiple objectives such as fairness or monotonicity? How can we get stable variable importance estimates? The Rashomon Effect provides surprising insight into all of these questions -- and more. 

The Rashomon Effect is often present in datasets  generated by processes that are nondeterministic, i.e., \textit{noisy} or \textit{uncertain}, including data used for bail and parole decisions, healthcare, and financial loan decisions -- high stakes applications. 
In fact, as we will discuss, it has been proven in specific cases that datasets drawn from noisy processes tend to exhibit a large Rashomon Effect  in that there are many approximately-equally accurate models \citep{SemenovaEtAl2023}.
Furthermore, a large Rashomon Effect correlates with the existence of simple-yet-accurate models \citep{SemenovaRuPa2022}. Hence, there is no accuracy/interpretability trade-off in many domains. The lack of a trade-off has been well-established empirically for tabular data \citep[e.g., see][]{holte1993very,lou2013accurate, AngelinoEtAl2018,mctavish2022fast, liu2022fast,McelfreshEtSAl2023}. This knowledge has important policy implications, because it explains that black box models have no performance advantage over interpretable models that are easier to administer and use.

Knowing that the Rashomon Effect exists -- and the extent to which it exists -- \textit{changes the lens through which we view just about everything in machine learning}. The current machine learning paradigm solves problems by finding a single good model. However, understanding that many good models differ dramatically -- in terms of variable importance, predictions on individual points, complexity, fairness, etc. -- changes how we approach the problem.
For instance, we cannot generally assume that any of the variables used heavily by one model are important to every well-performing model. We cannot assume that because an algorithm finds a complex model with good test performance that this level of complexity is necessarily needed for obtaining that test performance; similarly, we cannot assume that a complex model has discovered secrets in the dataset that a simpler model could not also find. Knowledge that many equally good models could exist might make us unhappy with the status quo of algorithms that optimize for only one machine learning model, when we could select from many. That is, just \textit{knowing about existence of the Rashomon Effect} shows us that the standard machine learning paradigm that returns only one model is too narrow, and new methods and insights are needed. 


We define the \textit{Rashomon set} as the set of models that perform approximately-equally to the best models from a given function class. 
The first algorithms that quantify the Rashomon Effect by capturing and storing the Rashomon set for nontrivial function classes have been developed only recently \citep{xin2022exploring,zhong2023exploring,LiuEtAlFasterRisk2022,zhu2023GroupFasterRisk}. These algorithms allow users to interact with the Rashomon set to address user preferences, such as fairness concerns and monotonicity constraints. They also allow us to study variable importance in a holistic way, including all of the good models.

We elaborate on how the Rashomon Effect has implications for simplicity, specifically the existence of simple-yet-accurate models (Section \ref{sec:covering}), flexibility to address user preferences without losing performance (Section \ref{sec:newparadigm}), uncertainty in predictions, models, and explanations (Section \ref{sec:multiplicity}), stable variable importance (Section \ref{sec:vi}), algorithm choice, specifically advance knowledge of which algorithms might be suitable for a given problem (Section \ref{sec:whichalg}), and public policy (Section \ref{sec:policy}).
We question the relevance of the classical machine learning paradigm in light of the Rashomon Effect in Section \ref{sec:narrow} and discuss an alternative paradigm in Section \ref{sec:newparadigm}.
This perspective piece distills work from several technical papers 
to make them more widely accessible and discusses their link to policy.

\section{The Rashomon Effect is Everywhere}
The Rashomon Effect occurs when there are many different well-performing models for the same dataset. Standard machine learning (ML) analysis reveals it, but most researchers might not recognize it, because they are not looking for it.

Let us work with a dataset -- the FICO dataset from the Explainable ML Challenge \citep{competition} -- though extremely similar results hold for an astounding number of other datasets \citep{SemenovaRuPa2022}.  We will examine the Rashomon Effect for a large model class, large enough to encompass the usual function classes used in machine learning, such as combinations of trees, neural networks, kernel machines, and so on. We applied a variety of machine learning methods to the data, including boosted decision trees, random forest, multi-layer perceptrons, support vector machines, logistic regression, and a 2-layer additive risk model. 
All of these models have completely different functional forms, from linear models to kernel-based nonparametric models with smooth decision boundaries, to tree-based nonparametric models with sharp decision boundaries, yet most of these models perform comparably, as shown in Table  \ref{tab: FICOTable}. This means all of these models are in the Rashomon set of the large class of functions.

\begin{table}[t]
\begin{tabular}{|l|l|l|}
\hline
Classifier & Test Accuracy & Test AUC \\ \hline
Random Forest & 0.697±0.017 & 0.757±0.017 \\ \hline
Boosted trees & 0.723±0.024   & 0.789±0.028 \\ \hline
SVM (linear kernel) & 0.720±0.029   & 0.795±0.027 \\ \hline
SVM (RBF kernel) & 0.727±0.023 & 0.799±0.022 \\ \hline
8-layer neural network & 0.722±0.022 & 0.792±0.026 \\ \hline
Logistic regression  & 0.731±0.023   & 0.801±0.028 \\ \hline
{\small 2-layer additive risk model} & 0.738±0.020 & 0.806±0.025 \\ \hline
\end{tabular}
\caption{Performance of different machine learning models on the 23-feature FICO dataset \citep{ChenFICO} over 10 test folds. They perform similarly. Some of these models (specifically, the 2-layer additive risk model) are interpretable.
}
\label{tab: FICOTable}
\end{table}

\begin{table*}[t]
\centering
\begin{tabular}{|l|l|l|l|}
\hline
Classifier & Reliance on & Reliance on & Reliance on  \\ 
& ExternalRiskEstimate & NumInqLast6M &  NetFractionRevolvingBurden \\\hline
Boosted trees &  1.18 ± 0.02 & 1.01 ± 0.00 & 1.03 ± 0.01 \\ \hline
SVM (RBF kernel) & 1.15 ± 0.01 &1.01 ± 0.00 & 1.12 ± 0.00 \\ \hline
Logistic regression  & 1.02 ± 0.00 & 1.02 ± 0.00& 1.23 ± 0.02\\ \hline
2-layer additive risk model & 1.01 ± 0.00 & 1.08 ± 0.01 & 1.00 ± 0.00\\ \hline
\end{tabular}
\caption{Model reliance, calculated by permuting feature values, \citep[the multiplicative version, from][]{fisher2019all} on three important features across top performing model classes, with standard error across 5 train-test splits. Here, e.g., 1.18 means the loss increases by 18\% when a variable is scrambled; in that case, the variable is quite important to the model. Value 1.00 means the variable is not important at all. The ordering of variable importance is inconsistent among these model classes; boosted trees rely most heavily on ``ExternalRiskEstimate,'' SVM relies heavily on both ``ExternalRiskEstimate'' and ``NetFractionRevolvingBurden,'' logistic regression relies most on ``NetFractionRevolingBurden,'' and the model from \citet{ChenFICO} relies most on ``NumInqLast6M.'' The variable ``NetFractionRevolingBurden'' is not important to the 2-layer additive risk models, but very important to logistic regression.}
\label{tab: FICO_MR}
\end{table*}

Table \ref{tab: FICO_MR} shows that these different models depend on variables differently. This exemplifies the Rashomon Effect -- when the best models that can be (practically) constructed for a given dataset are numerous and diverse. In Table \ref{tab: FICO_MR}, we used simple permutation importance \citep[e.g., see][]{fisher2019all} to estimate variable importance of each variable to the model's predictions.  This type of analysis can be conducted with similar results on many tabular datasets. 
We will see another way to directly observe the Rashomon Effect in Section \ref{sec:newparadigm}: by enumerating all of the good models from a given function class. Appendix \ref{app:metrics} shows other ways to measure the Rashomon Effect. 

Because there are many good functions, some of these functions are simple.

\section{The Rashomon Effect Gives Rise to Simpler-Yet-Accurate Models}\label{sec:covering}

\begin{figure}[ht]
    \centering
    \includegraphics[width = 0.8\columnwidth]{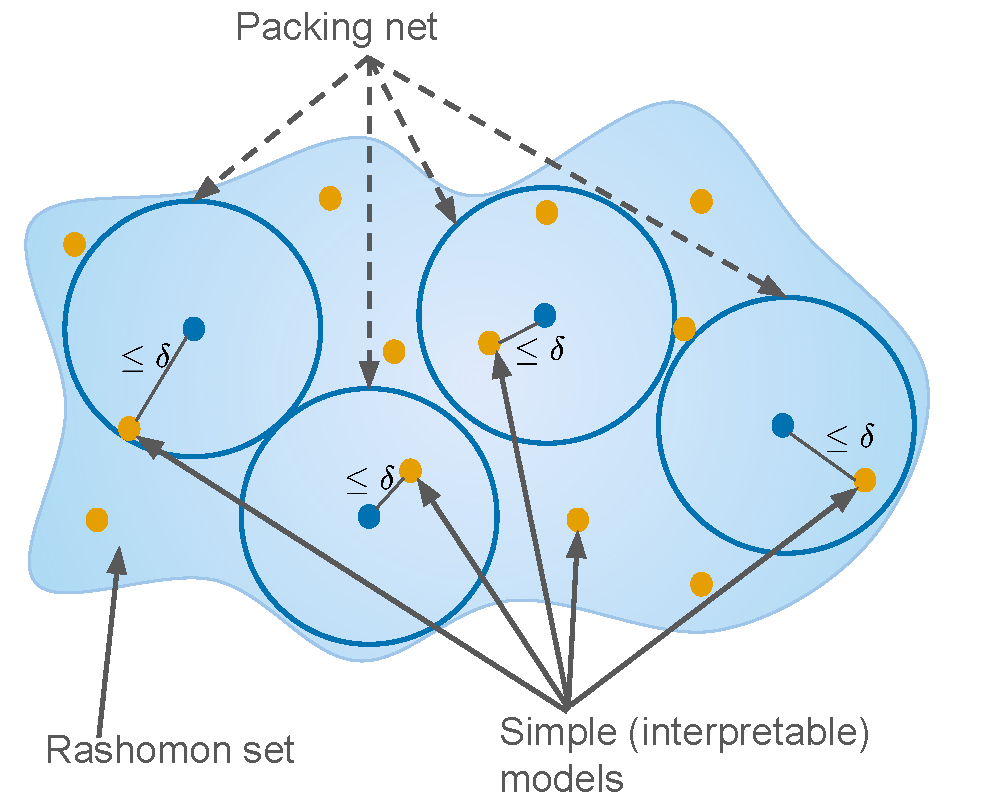}
    \caption{Illustration showing that for  hypothesis spaces with good approximating properties, larger Rashomon sets tend to contain multiple simpler models. For every model in the more complex space (blue shaded region), there exists a $\delta$-close model from the simpler space (orange dots). In this illustration, the Rashomon set contains at least four simpler models, which is its $2\delta$-packing number, where blue dots correspond to the centers of the balls in the packing.}
    \label{fig:rashomon_set_simplicity}
\end{figure}

When large portions of the function space contain many good models, the Rashomon set is likely large enough to include simpler models. A mathematical explanation for why this is true is illustrated in Figure \ref{fig:rashomon_set_simplicity}  \citep{SemenovaRuPa2022}. 
We have two function classes: a class of complex functions with a large Rashomon set (blue region) and a simpler function class contained in the complex function class (orange dots). 
Assume that every model in the complex class is ``close to'' a simpler one, meaning that they are within a radius $\delta$ of each other in function space (i.e., the simpler class is a \textit{cover} 
of the more complex set). Then, the Rashomon set in the complex class contains at least as many functions from the simpler class as its $2\delta$-packing number, where the packing number is the maximal number of balls in the Rashomon set, the centers of which are at least $2\delta$ apart. From the ``closeness'' assumption, every ball in the packing contains a simpler model, therefore, the larger the Rashomon set, the more simpler models it contains.

\begin{figure}[ht]
  \centering
\includegraphics[scale=0.5]{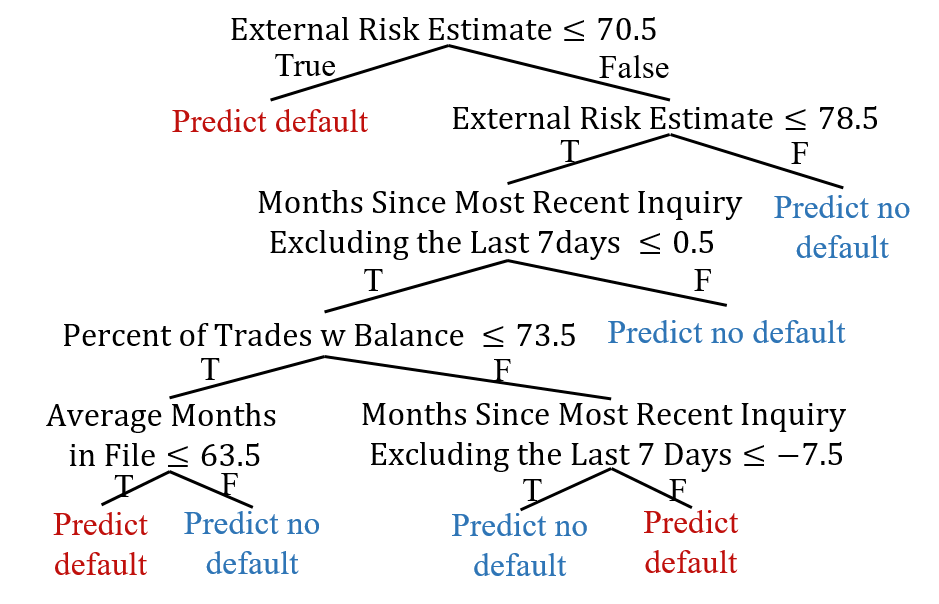}
\caption{Decision tree: train and test accuracy are both approximately 72\%, which is
comparable to the best black box algorithms (deep learning and boosted decision trees). This tree has 7 leaves and was obtained in 8.1 seconds by the GOSDT algorithm \citep{lin2020generalized, mctavish2022fast}.}
\label{fig: FICOTree}
\end{figure}

The ``closeness'' assumption is reasonable. For instance, sparse decision trees serve as a cover for deeper, more complex decision trees, and trees are universal function approximators \citep{barron1993universal}. If the more complex trees are optimized so that they are not so deep that they overfit, shallow trees serve as an even better cover.
For instance, Theorem 4.2 in \citet{mctavish2022fast} shows that any boosted decision tree is equivalent to a single tree with a greater depth, so the set of boosted trees is naturally covered by single trees. Because the closeness assumption generally holds, and because we often have large Rashomon sets, we often find that for tabular data, a single sparse tree (of depth $\sim$5) can achieve performance similar to that of a boosted decision tree. Figure \ref{fig: FICOTree} shows a single tree for the FICO dataset that achieves the accuracy of the black box models shown in Table \ref{tab: FICOTable}. Thus, as we stated, \textbf{for many problems, simple models can perform as well as much more complex models}, and there is no accuracy/interpretability trade-off.  

\begin{figure}[ht]
    \centering
    \includegraphics[width=0.95\linewidth]{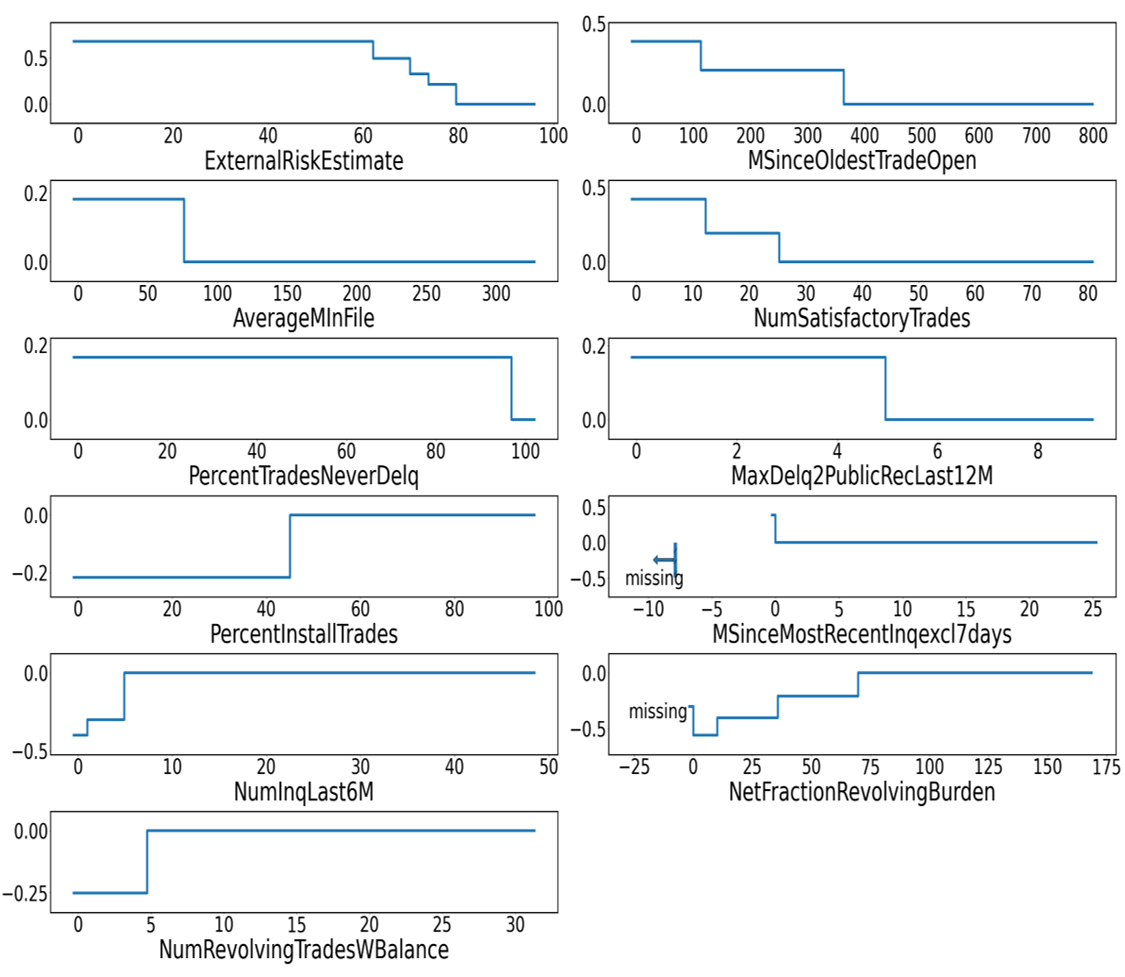}
    \caption{Sparse GAM: the user gets a score for each of the 11 features in this model, from the FastSparse algorithm \citep{liu2022fast}. The sum of scores translates into a risk of defaulting on a loan. The model was obtained in 3.15 seconds. Its test AUC and accuracy are .790 and .723. The cross-validation AUC and accuracy of  FastSparse are $0.791 \pm 0.010$ and $72.4\% \pm 1.2\%$.
    }
    \label{fig:FICOGAM}
\end{figure}

A second simple model, of a different functional form, is shown in Figure \ref{fig:FICOGAM}. This is a sparse generalized additive model (GAM). Interestingly, the GAM has no feature interaction terms yet achieves accuracy that is extremely similar to that of the decision trees, which rely only on feature interactions. Again, this illustrates a case of similar performance from completely different models.

This dataset was provided by FICO, the major credit scoring agency in the United States, for a data science competition to discover post-hoc explanations to black box models. It is a particularly interesting dataset, because the competition organizers thought interpretable-yet-accurate models did not exist, but they do. The dataset's variables represent credit history, and its label is whether an individual will default on a loan. The dataset appears to represent a particularly difficult (balanced) subpopulation that has a high risk of loan default relative to the population. If interpretable models demonstrate strong performance on this dataset, which was specifically designed for a benchmarking challenge, we could only imagine the potential success of interpretable models in addressing a much broader range of high-stakes problems.

Interpretable and simple models are easier to verify, easier to debug, and easier to use. However, they are not easier to design, as we discuss shortly. 


\section{The Standard ML Paradigm is Too Narrow}\label{sec:narrow}
Among models with equal complexity on a static dataset, statistical learning theory says that it should not matter which model in the Rashomon set we choose -- all models with equal complexity should generalize equally well to the test data \citep[e.g.,][chapter on learning theory]{Rudinbook}. This standard paradigm, where any model that has good cross-validation performance on a static dataset can be trusted, assumes the test data come from the same distribution as the training set, the data need no troubleshooting, and no additional domain knowledge is needed. This is why machine learning methods need only choose one model from the Rashomon set in the standard paradigm. 

However, the real world is not an ML benchmark.
In the real world, data are messy, the model inputs must be easy to check, domain knowledge can substitute for lack of data or problems with data, and models often need to be easy to understand and/or obey additional domain-specific constraints \citep{Wagstaff12,RudinWa14}. The standard ML paradigm makes it difficult to do just about anything except what it was designed for -- to produce accurate models on a static dataset.

Black box models have hidden flaws, whereas
users trying to design interpretable machine learning models realize quickly that \textbf{understandable models have understandable flaws}. This is a key reason why the standard machine learning paradigm often fails in the real world. Selecting just a single, understandable model -- ignoring the Rashomon Effect --   comes with problems. 
In our experience, the user is rarely satisfied with the first model that is output by a standard interpretable machine learning algorithm -- because they see flaws that can be fixed. Since standard ML algorithms are not interactive, the feedback loop can be fraught and frustrating. This is what we call the \textbf{interaction bottleneck}, where users cannot effectively interact with algorithms to improve machine learning models. Fortunately, there is a fix: finding Rashomon sets.

\section{A New Paradigm: Finding Rashomon Sets}\label{sec:newparadigm}
Because the Rashomon set is large, it often includes \textit{many} simple-yet-good models. Being able to \textbf{find all of the good models} from a given simple function class has benefits that we spend the rest of this article discussing, the most important ones appearing in this section. 

\begin{figure}[ht]
    \centering
    \includegraphics[width=0.9\linewidth]{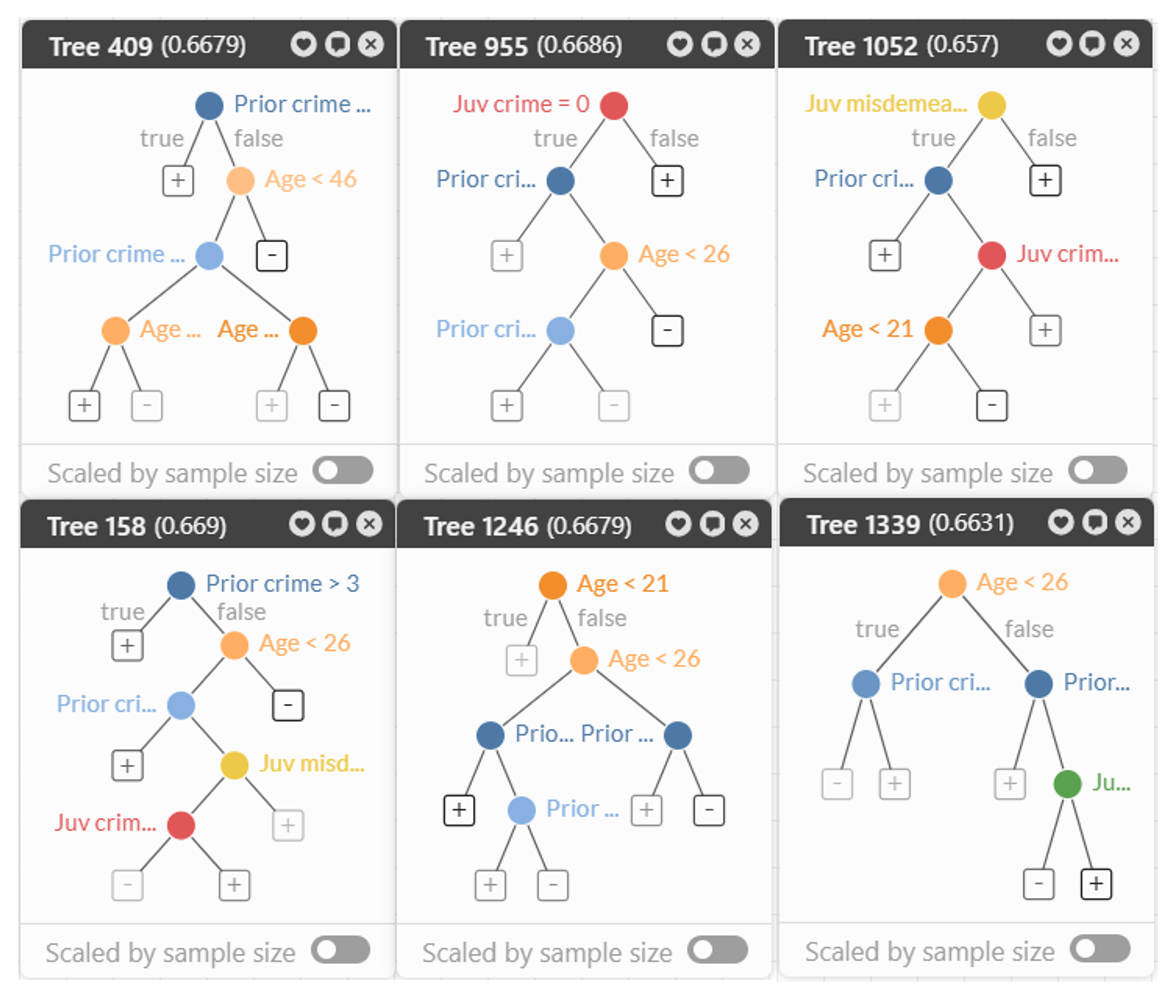}
    \caption{Example sparse decision trees in the Rashomon set of the COMPAS dataset \citep{LarsonMaKiAn16}, found by TreeFARMS. }
    \label{fig:DifferentTrees}
\end{figure}

The first algorithm that finds all good models for a nontrivial function class is the TreeFARMS algorithm, which finds all decision trees with a low regularized risk value \citep{xin2022exploring}. The second such algorithm is the GAM Rashomon set algorithm, which finds accurate sparse generalized additive models \citep{zhong2023exploring}. Third is the FasterRisk algorithm, which finds accurate sparse scoring systems \citep{LiuEtAlFasterRisk2022, zhu2023GroupFasterRisk}.
Figure \ref{fig:DifferentTrees} shows several sparse trees for the COMPAS dataset \citep{LarsonMaKiAn16} found using TreeFARMS.

The opportunity to search through the Rashomon set means we can optimize multiple objectives simultaneously, which has implications for
\textbf{constraint handling} and \textbf{alignment} with domain knowledge. Typical constraints that one might wish to include are \textit{monotonicity} constraints, where the predicted outcomes increase with a specified set of variables, and \textit{algorithmic fairness} constraints. 
A simple loop over the Rashomon set will suffice to find all possible answers to a constrained or multi-objective optimization problem.
Accordingly, the new paradigm allows us to easily align the model with multiple fairness objectives. \textbf{Among equally-good models, the user can choose one that optimally satisfies the criteria}.

\begin{figure}[ht]
    \centering
    \begin{subfigure}[b]{1\linewidth}
    \centering
    \includegraphics[width=0.95\linewidth]{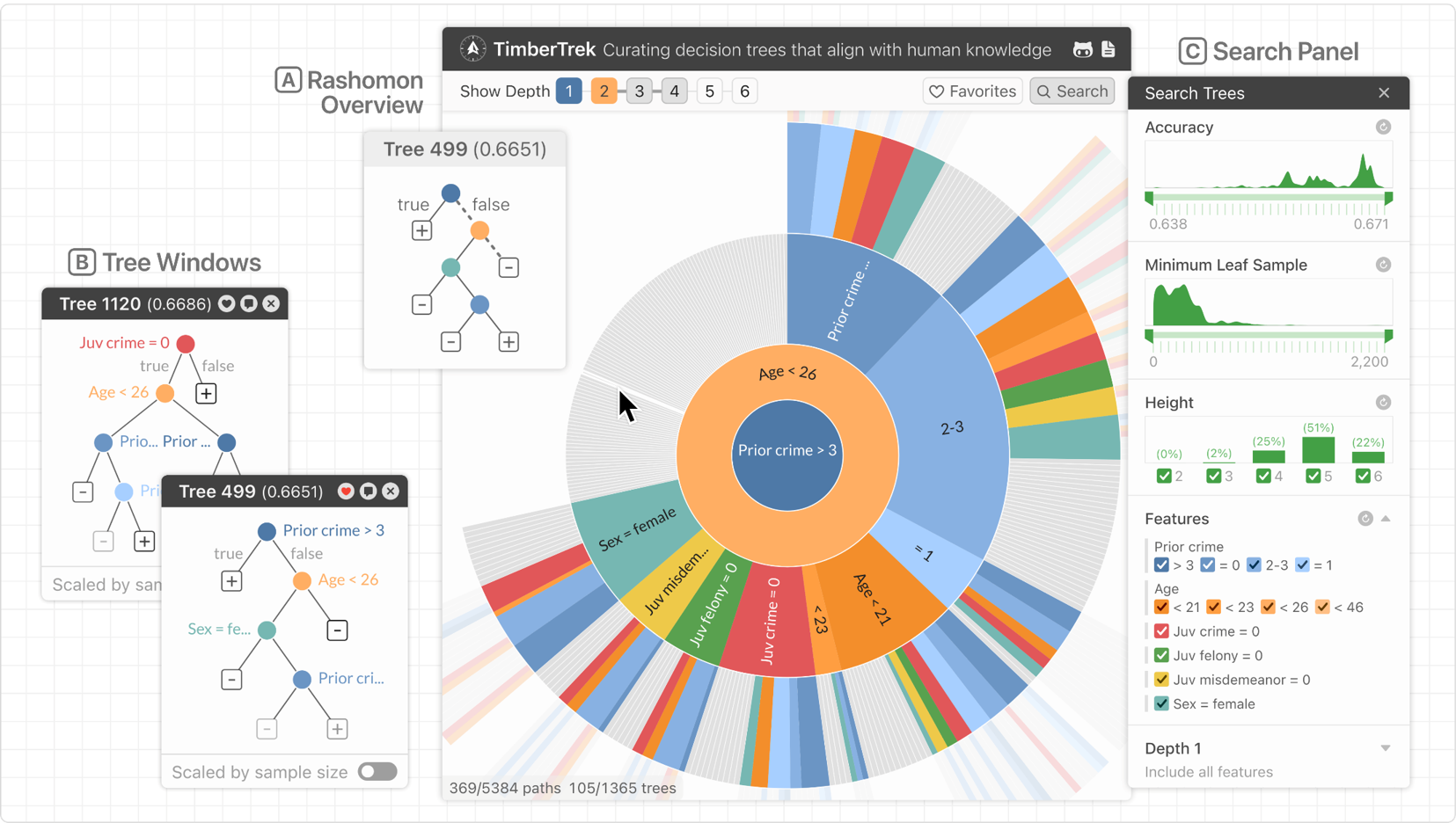}
    \caption{Rashomon set of the 1,365 best sparse decision trees for the COMPAS dataset, generated by TreeFARMS and displayed by TimberTrek  \citep[figure from][]{wang2022timbertrek}.}
    \label{fig:TimberTrek}
    \end{subfigure}
    \begin{subfigure}[b]{1\linewidth}
    \centering
    \includegraphics[width=0.95\linewidth]{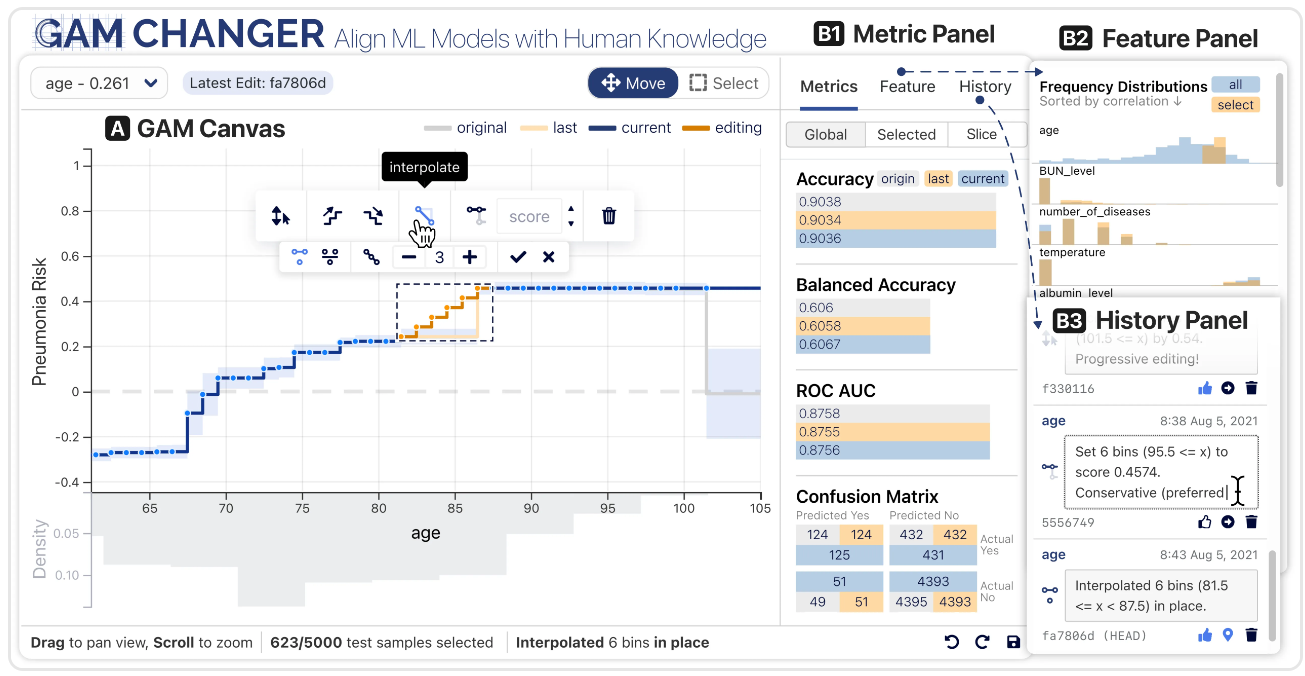}
    \caption{GAM Changer empowers domain experts and data scientists to easily and responsibly align model behaviors with their domain knowledge, via direct manipulation of GAM model weights \citep[figure from][]{wang2021gam}.}
    \label{fig:GAMChanger}
    \end{subfigure}
    \caption{Interactive tools. }
    \label{fig:InteractiveTools}
\end{figure}

Having the Rashomon set at one's fingertips \textbf{resolves the interaction bottleneck}. As discussed in the previous section, in the standard ML paradigm, if a user wants to add domain knowledge or constraints to the model, they need to formulate and solve new optimization problems each time they get new feedback from the user. This process is time-consuming, requires possibly many reformulations of optimization problems, and can be exceedingly frustrating. Access to the full Rashomon set resolves this. Using interactive tools, such as TimberTrek and GAMChanger \citep{wang2022timbertrek,wang2021gam},  to explore the Rashomon set, users can find a model that aligns with their domain knowledge in real time, even when that knowledge was not specified in advance. Figure \ref{fig:InteractiveTools} shows screenshots of these interactive tools. Even if the Rashomon set contains hundreds of millions of models, when they are organized effectively, humans can easily explore them.

The computational cost of finding the Rashomon set is much higher than that of finding a single optimal model. Luckily, the TreeFARMS, GAM Rashomon set, and FasterRisk algorithms can handle the  computation for most reasonably sized problems for sparse trees, generalized additive models, and scoring systems in minutes \citep[see timing tables in][]{xin2022exploring,zhong2023exploring,LiuEtAlFasterRisk2022}, which is often acceptable to users. If the Rashomon set is exceedingly large, these algorithms have mechanisms to sample from it or otherwise represent it. 

\textbf{Rashomon sets containing all accurate models contain Rashomon sets for a variety of other objectives}. Many of the objectives we consider in machine learning are related to each other. For instance, a highly accurate model probably also has high AUC, low loss, high F1-score, etc. We can take advantage of the relationship between these objectives. If we create a Rashomon set that includes all models with misclassification error below a threshold, we can often prove that it also includes all models with a \textit{different} objective below a different threshold. For example, \citet{xin2022exploring} showed how all models with high F1-score could be calculated without ever optimizing for F1-score directly. These models are easy to find, because they are contained in a high-accuracy Rashomon set.

Thus, Rashomon sets place substantially more control into the hands of human data analysts.


\section{Why Does the Rashomon Effect Occur?}\label{sec:noise}

\begin{figure*}[ht]
    \centering
    \includegraphics[width = \textwidth]{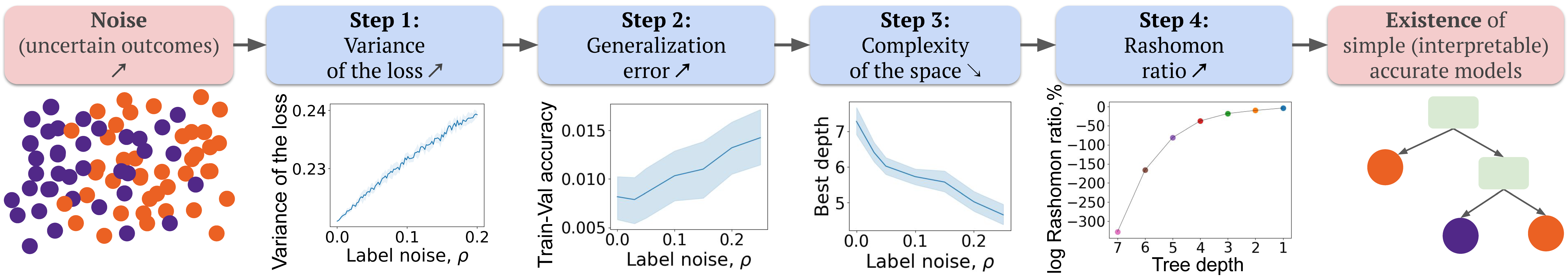}
    \caption{Description of the ``path'' that starts with noise and leads to larger Rashomon ratios and the existence of simpler models. The plots are created for the COMPAS dataset \cite{LarsonMaKiAn16}.}
    \label{fig:path}
\end{figure*}

One theory as to why the Rashomon Effect occurs for so many real-world datasets comes from ``A Path to Simpler Models Starts with Noise,'' by \citet{SemenovaEtAl2023}. This work shows that uncertainty in the outcomes (noise) is one of the causes of the Rashomon Effect. 
This work provides a 4-step ``path'' (see Figure \ref{fig:path}) that outlines how the uncertainty in the outcomes leads to simpler-yet-accurate models.

In Step 1 of the path, uncertainty in the outcomes (label noise) increases the variance of the loss function with respect to random draws of the data. This means the loss function's values are more uncertain, thus, the user cannot tell with certainty what the value of the loss might be on the test set by looking only at the training set. This leads to Step 2, where the generalization error, i.e., the difference between train and test performance, increases with the variance. Thus, 
it is easier to overfit the training set. 

In Step 3, the user realizes that they are overfitting, perhaps through conducting cross-validation, and simplifies the function class. Even stopping the path here, we arrive at simpler functions. The user had to simplify the function class, because they were not able to generalize, and the test performance would be poor if they did not do it. In other words, even if the data were generated from a complex process, as long as there is label noise, the user would still need to simplify the function class to avoid overfitting to that noise and to reduce cross-validation error.

In Step 4, when the function class is simpler, the \emph{Rashomon Ratio} is large. The Rashomon Ratio is the fraction of the model class that has close-to-optimal loss. It is the fraction of the function class within the Rashomon set. Thinking of these simpler functions as (diverse) representatives of the original function class, a large Rashomon Ratio for the simpler function class equates to a large number of well-performing representatives from the larger function class; i.e., a large Rashomon Effect. Thus, when the data generation process has noise, models with a wide range of complexity can all be part of the same large Rashomon set, leading to a large Rashomon Effect.


Tying this back to the existence of simpler-yet-accurate models from Section \ref{sec:covering}, 
the packing argument from Section \ref{sec:covering} can be used to show that models from an even simpler class are likely to exist that are approximately as accurate as functions within the user's current function class. 

A separate situation in which there is generally a large Rashomon Effect is when the margins between classes (distance between classes in feature space) are large. For example, the MNIST dataset is not a noisy dataset, yet almost any machine learning method -- with functions from any type of function class -- performs well on it.  

Although most real tabular datasets seem to have a large Rashomon Effect, it is easy to construct cases where Rashomon sets are extremely small. An example is where the labels are generated deterministically (no noise) from a complicated function. Incidentally, this is why \textit{approximating} or \textit{explaining} an \textit{already-selected} machine learning model will generally have an accuracy-interpretability trade-off, whereas working with real (noisy) labels will not. In other words, if we try to approximate a fixed, already-selected function $f$ with a simpler function $g$, then there will likely be a trade-off between the complexity of $g$ and how well it can fit $f$. This is discussed, for instance, by \citet{KleinbergMu20} (note that they reversed the terms ``interpretability'' and ``explainability'' from us).

\section{Uncertainty in Predictions, Fairness, and Explanations} \label{sec:multiplicity}

Knowledge of the Rashomon set can illuminate uncertainty that causes problems with ML systems. This includes \textit{underspecification} -- where the model development process does not have enough information to learn generalizable domain knowledge -- and \textit{predictive multiplicity}, where there are many different predictions made by models within the Rashomon set. If we can calculate the degree of predictive multiplicity in the Rashomon set (how many different predictions are possible), we gain insight into underspecification (many different conclusions).  Researchers have recently started to  quantify these effects \citep{marx2020predictive,coker2021theory,hsu2022rashomon,watson2023predictive}. 
Again, analysts typically minimize the \textit{loss} without considering the variation in other quantities, such as \textit{predictions}, \textit{variable importance}, or \textit{fairness}, which is where problems arise. 


\citet{d2020underspecification} demonstrate the impact of underspecification for several industry-scale learning tasks, such as medical imaging with eye and skin images, clinical risk prediction with electronic healthcare records, and pronoun affiliation with large language models; they observe that even the \textit{choice of random seed} can dramatically impact the behavior of the final model -- a serious consequence of overlooking the powerful Rashomon Effect.

Consider benchmarking challenges in algorithmic fairness. Here, one would use a special ``biased'' dataset, compute a model for it, find unfairness in this model, and fix the problem. However, \citet{cooper2023arbitrariness} points out that on these special datasets, if one averages over bootstrap samples to find a stable model, then such a model \textit{is already fair}. In other words, prior to \citet{cooper2023arbitrariness}, the improvement that fairness researchers were seeing could be a mirage due to the Rashomon Effect --  they were only considering one model from the Rashomon set that happened to be biased.  This follows the work of \citet{RodolfaLaGh21, CostonRaCh21, black2022model, blackLDA2024} showing that although there can exist an accuracy-fairness trade-off in theory \cite{kleinberg2018algfairness}, it may not exist in practice due to the Rashomon Effect. 



Rather than ignoring the Rashomon Effect, we should make use of it. 
We could, for instance, answer questions such as ``How  important can this variable possibly be for all good models?'' \citep{fisher2019all}, or ``Can I create a model with sparser counterfactual explanations?'' \citep{SunEtAl24}, or perhaps ``What is the largest and smallest my prediction could be from all of the good models?'' \citep{coker2021theory}. 
We can also visualize the Rashomon Effect by projecting the Rashomon set into variable importance space \citep{dong2020exploring}, where we can see how much every model in the Rashomon set depends on each variable. Or, we could leverage the Rashomon set to create stable variable importance values, discussed next.

\section{
Stable Variable Importance}\label{sec:vi}
Measuring the global significance of a variable in predicting an outcome holds paramount importance in scientific exploration and critical decision-making. Two important examples are genetics \citep[e.g.,][]{wang2020initial, novakovsky2022obtaining}, where the goal is to figure out which genes have unique information for predicting traits, and criminal justice, where mistakes in variable importance analysis have led to confusion and accusations of racial bias \citep[see][]{LarsonMaKiAn16, RudinWaCo2020}. 
Traditional approaches generally assess variable importance based on a \textit{single} model trained on a specific dataset, but this framework does not account for the Rashomon Effect. 
As we know, failing to consider it can lead different researchers to draw divergent conclusions from identical data, based on identifying different variables as important. Along with the Rashomon Effect, another issue in variable importance is the lack of reproducibility: a variable importance estimate can change amid reasonable data perturbations (e.g., swapping out on observation).
One solution is provided by the Rashomon Importance Distribution (RID), which quantifies the importance of a variable across the set of all good models and across perturbations to the original dataset using almost any variable importance metric of interest \citep{donnelly2023the}. By considering variations of the data through bootstrapping, and considering the Rashomon set for each bootstrap sample to produce a distribution of variable importance values for each variable, RID can obtain much more stable variable importance calculations. It is able to recover variables that are important to complex data generation processes more accurately than other approaches, demonstrating how leveraging knowledge of the Rashomon Effect can be helpful for scientific discovery.

\section{Which Algorithm Should I Use?}\label{sec:whichalg}
A perennial question in machine learning is about the match of algorithms to problems: which machine learning algorithm is likely to work for my data? For image and text data there are clear current answers that take advantage of the structure in these data types (e.g., CNNs and transformers), but for tabular data there is not -- most machine learning algorithms perform equally well.  \citep[In fact, researchers have had to compile special ``hard'' datasets because it is uncommon to find cases where different ML algorithms perform differently, see][]{McelfreshEtSAl2023}.
From what we have discussed above, the answer depends on the level of noise in the outcomes. 

For predicting criminal recidivism, where we predict months or years in advance whether someone will commit a crime, the randomness in this process means that simpler models will tend to perform as well as complex models. Thus, we would expect that boosted decision trees \citep{Freund95}, random forest \citep{BreimanRF}, or neural networks provide no performance advantage over sparse additive models or the type of simple scoring systems that are often used for this purpose by the criminal justice system. Empirical evidence on recidivism prediction supports this \citep[see e.g.,][]{WangHanEtAl2022,ZengUsRu2017, tollenaar2013method}. Results of this flavor would be expected to hold for loan default predictions, and we presented empirical evidence based on the FICO dataset earlier. This same reasoning process holds for many healthcare prediction problems \citep[readmission, mortality, e.g., see][]{zhu2023GroupFasterRisk}. In other words, simply by knowing the type of data and the level of noise in the outcome, we can determine whether methods that produce optimized simpler models are likely to be sufficiently accurate. 

It is important to note that there is a performance difference between algorithms such as CART \citep{BreimanCART} and C4.5 \citep{Quinlan} from the 1980s and 1990s and more modern algorithms. As shown by \citet{xin2022exploring}, CART rarely produces models within the Rashomon set of a dataset, even when subsampling data numerous times and rerunning CART for each subsample. This means these older algorithms do not achieve an accuracy/sparsity balance that is as good as more modern algorithms.  \citep[Additional experiments for CART vs$.$ more modern tree algorithms appear in][]{lin2020generalized,mctavish2022fast}. 

Thus, in terms of best practices for standard (``noisy'') tabular data in cases where interpretability is important, one would typically first \textit{find baseline performance using black box models.}
Then, one would \textit{try to match baseline performance using modern interpretable modeling algorithms} (from the 2020s rather than the 1980s). The easiest interpretable ML algorithm to start with is probably FastSparse \citep{liu2022fast}, which produces sparse generalized additive models. 
Decision trees are a much harder class to optimize, so when working with them, it is useful to use a reduction in the search space provided by GOSDT+Guesses \cite{mctavish2022fast,lin2020generalized}; here, the splits from boosted decision trees are used for constructing single high-accuracy trees. There are quite a lot of interpretable ML algorithms available  that produce models of a variety of functional forms. Scoring systems \citep[e.g., as produced by the FasterRisk algorithm of][]{LiuEtAlFasterRisk2022}, are extremely popular in medicine and criminal justice. For linear models, OKRidge~\citep{liu2023okridge} can find sparse solutions with provable optimality. Rule sets \citep[e.g.,][]{WangEtAl2017} are simple logical models. For datasets requiring more complex models, try GA2M models \citep{lou2013accurate}, which are additive models with pairwise interactions. Or, as is typically done in credit risk scoring,  one could create several smaller models (subscales), and combine them using a small model, as in the 2-layer additive risk model from \citet{ChenFICO} that was used in Table \ref{tab: FICOTable}. Usually each subscale represents a category of features (e.g., credit delinquency features, or satisfactory trade features). 


Finally, assuming these ``first try'' interpretable models are flawed in some way that a user can identify, we suggest \textit{allowing the user to explore the Rashomon set using an algorithm and interface} such as TreeFARMS \& TimberTrek or GAM Rashomon set \& GAM Changer. This should yield a model suitable for further consideration.

\section{Policy Implications}\label{sec:policy}
Knowledge of the Rashomon Effect can be used to deliver significant positive impacts to society, including the development of fairer and more interpretable models. 

Currently, policy makers have started to govern the ``right to explanation'' for certain algorithmic decisions \citep{RightToExpl}. However, companies often do not want to provide models that could provide an advantage to competitors. This tension between a desire to preserve secrecy and mandated explanations leads to them providing narrow explanations that can be both misleading and incomplete, rather then genuinely transparent. Explanations are generally \textit{post hoc}, which introduces several possible problems. First, they might be unfaithful to the underlying reasoning process, e.g., ``You were denied a loan due to factors A and B,'' when, in fact, the loan denial was due to different factors. Second, the explanations might be so incomplete as to be practically useless, e.g., ``Factors A and B are important in our decision,'' with no further explanation of how they were used and whether other factors might also be important. A person receiving an explanation has no way to determine the quality of that explanation. 
Problems with explanations have been discussed at length \citep[e.g.,][]{AdebayoEtAl2018,Rudin19,YanagawaEtAl2024, han2022explanation}. Essentially, black box models, even when supplemented with explanations, create barriers for individuals to examine and question the models, effectively allowing model designers to hide their flaws. 

Interpretable models do not have any of these issues. Their explanations must be faithful and complete by design. They are much easier to troubleshoot and use in practice. And, as we discussed in Sections \ref{sec:covering} and \ref{sec:noise}, the Rashomon Effect theoretically explains why and when interpretable models perform as well as their black box counterparts. 
For these reasons, \textbf{interpretable models should be used by default for many high-stakes decisions using machine learning}.  Thus, for applications such as criminal recidivism, we should default to interpretable models when we know the outcomes are noisy and where empirical evidence on similar problems has confirmed that interpretable models perform well \citep[see, e.g.,][]{WangHanEtAl2022, ZengUsRu2017}.
Exceptions can be made in cases where models are 100\% accurate (e.g., lesion detection in medical images), in cases where no reason is needed (e.g., medical image segmentation), or in cases where there is no practical way to create an interpretable model. However, since we can now find the Rashomon set, as discussed in Section \ref{sec:newparadigm}, making it easier to build interpretable models, there is often no excuse to continue the use of black boxes.

Another policy implication involves other types of fairness besides simplicity. We can find the ``most fair'' model within the Rashomon set, according to any fairness metric, including recourse \citep[e.g.,][]{black2022model}, and thus \textbf{can verify claims about whether there exists a fairer-yet-accurate model for a given dataset.}

Even though interpretability, uncertainty, and fairness are essential to AI in practice and policy -- with the Rashomon Effect being central to all of them -- these topics are touched upon only superficially in most of today's academic courses. With respect to interpretability, most courses introduce no techniques more modern than CART \citep{BreimanCART} and C4.5 \citep{Quinlan}.
Information on post hoc explanations is much more widespread, sometimes (unfortunately) using the terminology ``interpretable'' to describe them. A review of interpretable machine learning appears in \citet{rudin2022interpretable}, and course material is available at \citet{Rudinbook}. 
Policy makers can fund \textbf{ethical AI education, which will inevitably involve the Rashomon Effect} since it determines whether trade-offs can exist between performance and ethical AI objectives.

\section{Conclusion}
The Rashomon Effect shows us that among models with similar \textit{loss}, there are a multitude of models with different \textit{properties}, including various levels of simplicity, fairness, and explanations/variable importance values. 

The ability to capture Rashomon sets and display them to users addresses what is arguably the hardest open problem in interpretable machine learning -- incorporating human interaction.
Solving the interaction bottleneck can have a major impact on our ability to troubleshoot and add constraints, which, in turn, could have a major impact on whether machine learning models can be used in high-stakes decisions. 

We do not believe that we have truly grasped the full extent of the Rashomon Effect yet, but we can already see that its impact on practical machine learning will be enormous. It forces us to change the way we think -- even back to the fundamentals of ML. Since we formulate ML algorithms in terms of trade-offs between objectives, we tend to think that trade-offs among these objectives must then exist in the models they create. This is -- surprisingly -- wrong.





\section*{Acknowledgements}
We acknowledge support from DOE DE-SC0023194, NSF IIS-2130250, NIDA R01 DA054994, NSF IIS-2147061, and a grant from Fujitsu. We also acknowledge the support of the Natural Sciences and Engineering Research Council of Canada (NSERC). Nous remercions le Conseil de recherches en sciences naturelles et en génie du Canada (CRSNG) de son soutien.

\section*{Impact Statement}
There are significant societal impacts discussed in this work, with the most important points summarized as: 
(1) Rashomon sets often admit many good models, giving rise to the existence of high-performing models that obey constraints such as interpretability and fairness; such constraints are crucial in high-stakes settings. 
(2) Machine learning algorithms now exist within the new Rashomon set paradigm. These algorithms can find whole Rashomon sets for a given dataset, mitigating the \textit{interaction bottleneck}, and allowing users to easily create usable machine learning models for a huge variety of applications. 
(3) We can determine, before seeing any data, and by knowing only that noise is present in the data generation process, whether a large Rashomon set will exist, and (thus) whether simpler and/or fairer well-performing models will exist. Policy-makers can use this information as evidence for mandating that interpretable models be used for many high-stakes decisions by default. In this way, knowledge of the Rashomon set and its origins can help make the practical uses of machine learning safer and fairer across society.
(4) \textit{Knowledge} of the Rashomon Effect changes the way we view just about everything in machine learning, including uncertainty, variable importance measurements, interpretability, fairness, interactivity, and even the classical paradigm of machine learning.

\bibliography{refs}

\begin{thebibliography}{61}
\providecommand{\natexlab}[1]{#1}
\providecommand{\url}[1]{\texttt{#1}}
\expandafter\ifx\csname urlstyle\endcsname\relax
  \providecommand{\doi}[1]{doi: #1}\else
  \providecommand{\doi}{doi: \begingroup \urlstyle{rm}\Url}\fi

\bibitem[Adebayo et~al.(2018)Adebayo, Gilmer, Muelly, Goodfellow, Hardt, and
  Kim]{AdebayoEtAl2018}
Adebayo, J., Gilmer, J., Muelly, M., Goodfellow, I., Hardt, M., and Kim, B.
\newblock Sanity checks for saliency maps.
\newblock In Bengio, S., Wallach, H., Larochelle, H., Grauman, K.,
  Cesa-Bianchi, N., and Garnett, R. (eds.), \emph{Neural Information Processing
  Systems ({NeurIPS})}, volume~31, 2018.

\bibitem[Ahanor et~al.(2022)Ahanor, Medal, and Trapp]{ahanor2022diversitree}
Ahanor, I., Medal, H., and Trapp, A.~C.
\newblock Diversitree: Computing diverse sets of near-optimal solutions to
  mixed-integer optimization problems.
\newblock \emph{arXiv preprint arXiv:2204.03822}, 2022.

\bibitem[Angelino et~al.(2018)Angelino, Larus-Stone, Alabi, Seltzer, and
  Rudin]{AngelinoEtAl2018}
Angelino, E., Larus-Stone, N., Alabi, D., Seltzer, M., and Rudin, C.
\newblock Learning certifiably optimal rule lists for categorical data.
\newblock \emph{Journal of Machine Learning Research}, 18\penalty0
  (234):\penalty0 1--78, 2018.

\bibitem[Barron(1993)]{barron1993universal}
Barron, A.~R.
\newblock Universal approximation bounds for superpositions of a sigmoidal
  function.
\newblock \emph{IEEE Transactions on Information Theory}, 39\penalty0
  (3):\penalty0 930--945, 1993.

\bibitem[Black et~al.(2022)Black, Raghavan, and Barocas]{black2022model}
Black, E., Raghavan, M., and Barocas, S.
\newblock Model multiplicity: Opportunities, concerns, and solutions.
\newblock In \emph{2022 ACM Conference on Fairness, Accountability, and
  Transparency}, pp.\  850--863, 2022.

\bibitem[Black et~al.(2024)Black, Koepke, Kim, Barocas, and Hsu]{blackLDA2024}
Black, E., Koepke, J.~L., Kim, P., Barocas, S., and Hsu, M.
\newblock Less discriminatory algorithms.
\newblock \emph{Georgetown Law Journal}, 113\penalty0 (1), 2024.
\newblock Washington University in St Louis Legal Studies Research Paper
  (forthcoming).

\bibitem[Breiman(2001{\natexlab{a}})]{BreimanRF}
Breiman, L.
\newblock Random {Forests}.
\newblock \emph{Machine Learning}, 45\penalty0 (1):\penalty0 5--32,
  2001{\natexlab{a}}.

\bibitem[Breiman(2001{\natexlab{b}})]{breiman2001statistical}
Breiman, L.
\newblock Statistical modeling: The two cultures (with comments and a rejoinder
  by the author).
\newblock \emph{Statistical Science}, 16\penalty0 (3):\penalty0 199--231,
  2001{\natexlab{b}}.

\bibitem[Breiman et~al.(1984)Breiman, Friedman, Stone, and Olshen]{BreimanCART}
Breiman, L., Friedman, J., Stone, C.~J., and Olshen, R.~A.
\newblock \emph{Classification and {Regression} {Trees}}.
\newblock CRC press, 1984.

\bibitem[Chen et~al.(2022)Chen, Lin, Rudin, Shaposhnik, Wang, and
  Wang]{ChenFICO}
Chen, C., Lin, K., Rudin, C., Shaposhnik, Y., Wang, S., and Wang, T.
\newblock A holistic approach to interpretability in financial lending: Models,
  visualizations, and summary-explanations.
\newblock \emph{Decision Support Systems}, 152:\penalty0 113647, 2022.

\bibitem[Coker et~al.(2021)Coker, Rudin, and King]{coker2021theory}
Coker, B., Rudin, C., and King, G.
\newblock A theory of statistical inference for ensuring the robustness of
  scientific results.
\newblock \emph{Management Science}, 67\penalty0 (10):\penalty0 6174--6197,
  2021.

\bibitem[Cooper et~al.(2024)Cooper, Lee, Choksi, Barocas, Sa, Grimmelmann,
  Kleinberg, Sen, and Zhang]{cooper2023arbitrariness}
Cooper, A.~F., Lee, K., Choksi, M., Barocas, S., Sa, C.~D., Grimmelmann, J.,
  Kleinberg, J., Sen, S., and Zhang, B.
\newblock Arbitrariness and prediction: The confounding role of variance in
  fair classification.
\newblock In \emph{Proceedings of the {AAAI} Conference on Artificial
  Intelligence}, volume~38, pp.\  22004--22012, Mar. 2024.

\bibitem[Coston et~al.(2021)Coston, Rambachan, and Chouldechova]{CostonRaCh21}
Coston, A., Rambachan, A., and Chouldechova, A.
\newblock Characterizing fairness over the set of good models under selective
  labels.
\newblock In \emph{Proceedings of the International Conference on Machine
  Learning (ICML)}, volume 139, pp.\  2144--2155, 18--24 Jul 2021.

\bibitem[Dong \& Rudin(2020)Dong and Rudin]{dong2020exploring}
Dong, J. and Rudin, C.
\newblock Exploring the cloud of variable importance for the set of all good
  models.
\newblock \emph{Nature Machine Intelligence}, 2\penalty0 (12):\penalty0
  810--824, 2020.

\bibitem[Donnelly et~al.(2023)Donnelly, Katta, Rudin, and
  Browne]{donnelly2023the}
Donnelly, J., Katta, S., Rudin, C., and Browne, E.~P.
\newblock The rashomon importance distribution: Getting {RID} of unstable,
  single model-based variable importance.
\newblock In \emph{Neural Information Processing Systems ({NeurIPS})}, 2023.

\bibitem[D’Amour et~al.(2020)D’Amour, Heller, Moldovan, Adlam, Alipanahi,
  Beutel, Chen, Deaton, Eisenstein, Hoffman, et~al.]{d2020underspecification}
D’Amour, A., Heller, K., Moldovan, D., Adlam, B., Alipanahi, B., Beutel, A.,
  Chen, C., Deaton, J., Eisenstein, J., Hoffman, M.~D., et~al.
\newblock Underspecification presents challenges for credibility in modern
  machine learning.
\newblock \emph{Journal of Machine Learning Research}, 2020.

\bibitem[{FICO} et~al.(2018){FICO}, {Google}, {Imperial College London}, {MIT},
  {University of Oxford}, {UC Irvine}, and {UC Berkeley}]{competition}
{FICO}, {Google}, {Imperial College London}, {MIT}, {University of Oxford}, {UC
  Irvine}, and {UC Berkeley}.
\newblock {Explainable Machine Learning Challenge}.
\newblock
  \url{https://community.fico.com/s/explainable-machine-learning-challenge},
  2018.

\bibitem[Fisher et~al.(2019)Fisher, Rudin, and Dominici]{fisher2019all}
Fisher, A., Rudin, C., and Dominici, F.
\newblock All models are wrong, but many are useful: Learning a variable's
  importance by studying an entire class of prediction models simultaneously.
\newblock \emph{Journal of Machine Learning Research}, 20\penalty0
  (177):\penalty0 1--81, 2019.

\bibitem[Freund \& Schapire(1997)Freund and Schapire]{Freund95}
Freund, Y. and Schapire, R.
\newblock A decision-theoretic generalization of on-line learning and an
  application to boosting.
\newblock \emph{Journal of Computer and System Sciences}, 55\penalty0
  (1):\penalty0 119--139, August 1997.

\bibitem[Han et~al.(2022)Han, Srinivas, and Lakkaraju]{han2022explanation}
Han, T., Srinivas, S., and Lakkaraju, H.
\newblock Which explanation should {I} choose? a function approximation
  perspective to characterizing post hoc explanations.
\newblock In \emph{Neural Information Processing Systems ({NeurIPS})},
  volume~35, pp.\  5256--5268, 2022.

\bibitem[Holte(1993)]{holte1993very}
Holte, R.~C.
\newblock Very simple classification rules perform well on most commonly used
  datasets.
\newblock \emph{Machine Learning}, 11:\penalty0 63--90, 1993.

\bibitem[Hsu \& Calmon(2022)Hsu and Calmon]{hsu2022rashomon}
Hsu, H. and Calmon, F.
\newblock Rashomon capacity: A metric for predictive multiplicity in
  classification.
\newblock In \emph{Neural Information Processing Systems ({NeurIPS})},
  volume~35, pp.\  28988--29000, 2022.

\bibitem[Kleinberg(2018)]{kleinberg2018algfairness}
Kleinberg, J.
\newblock Inherent trade-offs in algorithmic fairness.
\newblock \emph{SIGMETRICS Perform. Eval. Rev.}, 46\penalty0 (1):\penalty0 40,
  June 2018.

\bibitem[Kleinberg \& Mullainathan(2019)Kleinberg and
  Mullainathan]{KleinbergMu20}
Kleinberg, J. and Mullainathan, S.
\newblock Simplicity creates inequity: Implications for fairness, stereotypes,
  and interpretability.
\newblock In \emph{Proceedings of the 2019 ACM Conference on Economics and
  Computation}, EC '19, pp.\  807–808, 2019.

\bibitem[Kurosawa(1950)]{RashMovie}
Kurosawa, A.
\newblock Rashomon.
\newblock RKO Radio Pictures, 1950.

\bibitem[Larson et~al.(2016)Larson, Mattu, Kirchner, and
  Angwin]{LarsonMaKiAn16}
Larson, J., Mattu, S., Kirchner, L., and Angwin, J.
\newblock How we analyzed the {COMPAS} recidivism algorithm.
\newblock \emph{ProPublica}, 2016.

\bibitem[Lin et~al.(2020)Lin, Zhong, Hu, Rudin, and
  Seltzer]{lin2020generalized}
Lin, J., Zhong, C., Hu, D., Rudin, C., and Seltzer, M.
\newblock Generalized and scalable optimal sparse decision trees.
\newblock In \emph{Proceedings of the International Conference on Machine
  Learning (ICML)}, pp.\  6150--6160, 2020.

\bibitem[Liu et~al.(2022{\natexlab{a}})Liu, Zhong, Li, Seltzer, and
  Rudin]{LiuEtAlFasterRisk2022}
Liu, J., Zhong, C., Li, B., Seltzer, M., and Rudin, C.
\newblock Fasterrisk: Fast and accurate interpretable risk scores.
\newblock In \emph{Neural Information Processing Systems ({NeurIPS})},
  2022{\natexlab{a}}.

\bibitem[Liu et~al.(2022{\natexlab{b}})Liu, Zhong, Seltzer, and
  Rudin]{liu2022fast}
Liu, J., Zhong, C., Seltzer, M., and Rudin, C.
\newblock Fast sparse classification for generalized linear and additive
  models.
\newblock In \emph{Proceedings of Artificial Intelligence and Statistics
  (AISTATS)}, 2022{\natexlab{b}}.

\bibitem[Liu et~al.(2023)Liu, Rosen, Zhong, and Rudin]{liu2023okridge}
Liu, J., Rosen, S., Zhong, C., and Rudin, C.
\newblock O{KR}idge: Scalable optimal k-sparse ridge regression.
\newblock In \emph{Neural Information Processing Systems ({NeurIPS})}, 2023.

\bibitem[Lou et~al.(2013)Lou, Caruana, Gehrke, and Hooker]{lou2013accurate}
Lou, Y., Caruana, R., Gehrke, J., and Hooker, G.
\newblock Accurate intelligible models with pairwise interactions.
\newblock In \emph{Proceedings of the 19th ACM SIGKDD International Conference
  on Knowledge Discovery and Data Mining}, pp.\  623--631, 2013.

\bibitem[Marx et~al.(2020)Marx, Calmon, and Ustun]{marx2020predictive}
Marx, C., Calmon, F., and Ustun, B.
\newblock Predictive multiplicity in classification.
\newblock In \emph{Proceedings of the International Conference on Machine
  Learning (ICML)}, pp.\  6765--6774, 2020.

\bibitem[Mata et~al.(2022)Mata, Kanamori, and Arimura]{mata2022computing}
Mata, K., Kanamori, K., and Arimura, H.
\newblock Computing the collection of good models for rule lists.
\newblock \emph{arXiv preprint arXiv:2204.11285}, 2022.

\bibitem[McElfresh et~al.(2023)McElfresh, Khandagale, Valverde, C, Feuer,
  Hegde, Ramakrishnan, Goldblum, and White]{McelfreshEtSAl2023}
McElfresh, D., Khandagale, S., Valverde, J., C, V.~P., Feuer, B., Hegde, C.,
  Ramakrishnan, G., Goldblum, M., and White, C.
\newblock When do neural nets outperform boosted trees on tabular data?
\newblock In \emph{Neural Information Processing Systems ({NeurIPS})}, 2023.

\bibitem[McTavish et~al.(2022)McTavish, Zhong, Achermann, Karimalis, Chen,
  Rudin, and Seltzer]{mctavish2022fast}
McTavish, H., Zhong, C., Achermann, R., Karimalis, I., Chen, J., Rudin, C., and
  Seltzer, M.
\newblock Fast sparse decision tree optimization via reference ensembles.
\newblock In \emph{Proceedings of the AAAI Conference on Artificial
  Intelligence}, volume~36, pp.\  9604--9613, 2022.

\bibitem[Novakovsky et~al.(2022)Novakovsky, Dexter, Libbrecht, Wasserman, and
  Mostafavi]{novakovsky2022obtaining}
Novakovsky, G., Dexter, N., Libbrecht, M.~W., Wasserman, W.~W., and Mostafavi,
  S.
\newblock Obtaining genetics insights from deep learning via explainable
  artificial intelligence.
\newblock \emph{Nature Reviews Genetics}, pp.\  1--13, 2022.

\bibitem[Quinlan(1993)]{Quinlan}
Quinlan, J.~R.
\newblock \emph{C4.5: programs for machine learning}, volume~1.
\newblock Morgan Kaufmann, 1993.

\bibitem[Rodolfa et~al.(2021)Rodolfa, Lamba, and Ghani]{RodolfaLaGh21}
Rodolfa, K.~T., Lamba, H., and Ghani, R.
\newblock Empirical observation of negligible fairness–accuracy trade-offs in
  machine learning for public policy.
\newblock \emph{Nature Machine Intelligence}, 3:\penalty0 896--–904, October
  2021.

\bibitem[Rudin(2019)]{Rudin19}
Rudin, C.
\newblock Stop explaining black box machine learning models for high stakes
  decisions and use interpretable models instead.
\newblock \emph{Nature Machine Intelligence}, 1:\penalty0 206--215, May 2019.

\bibitem[Rudin(2020)]{Rudinbook}
Rudin, C.
\newblock \emph{Intuition for the Algorithms of Machine Learning}.
\newblock self-published at
  \url{https://users.cs.duke.edu/~cynthia/teaching.html}, 2020.

\bibitem[Rudin \& Wagstaff(2014)Rudin and Wagstaff]{RudinWa14}
Rudin, C. and Wagstaff, K.~L.
\newblock Machine learning for science and society.
\newblock \emph{Machine Learning}, 95\penalty0 (1), 2014.

\bibitem[Rudin et~al.(2020)Rudin, Wang, and Coker]{RudinWaCo2020}
Rudin, C., Wang, C., and Coker, B.
\newblock The \textit{Age} of secrecy and unfairness in recidivism prediction.
\newblock \emph{Harvard Data Science Review}, 2\penalty0 (1), 1 2020.

\bibitem[Rudin et~al.(2022)Rudin, Chen, Chen, Huang, Semenova, and
  Zhong]{rudin2022interpretable}
Rudin, C., Chen, C., Chen, Z., Huang, H., Semenova, L., and Zhong, C.
\newblock Interpretable machine learning: Fundamental principles and 10 grand
  challenges.
\newblock \emph{Statistics Surveys}, 16:\penalty0 1--85, 2022.

\bibitem[Semenova et~al.(2022)Semenova, Rudin, and Parr]{SemenovaRuPa2022}
Semenova, L., Rudin, C., and Parr, R.
\newblock On the existence of simpler machine learning models.
\newblock In \emph{{ACM} Conference on Fairness, Accountability, and
  Transparency ({ACM FAccT})}, 2022.

\bibitem[Semenova et~al.(2023)Semenova, Chen, Parr, and
  Rudin]{SemenovaEtAl2023}
Semenova, L., Chen, H., Parr, R., and Rudin, C.
\newblock A path to simpler models starts with noise.
\newblock In \emph{Neural Information Processing Systems ({NeurIPS})}, 2023.

\bibitem[Smith et~al.(2020)Smith, Mansilla, and Goulding]{smith2020model}
Smith, G., Mansilla, R., and Goulding, J.
\newblock Model class reliance for random forests.
\newblock In \emph{Neural Information Processing Systems ({NeurIPS})},
  volume~33, pp.\  22305--22315, 2020.

\bibitem[Sun et~al.(2024)Sun, Chen, Orlandi, Wang, and Rudin]{SunEtAl24}
Sun, Y., Chen, Z., Orlandi, V., Wang, T., and Rudin, C.
\newblock Sparse and faithful explanations without sparse models.
\newblock In \emph{Proc. Artificial Intelligence and Statistics {(AISTATS)}},
  2024.

\bibitem[Tollenaar \& {van der Heijden}(2013)Tollenaar and {van der
  Heijden}]{tollenaar2013method}
Tollenaar, N. and {van der Heijden}, P.
\newblock Which method predicts recidivism best?: A comparison of statistical,
  machine learning and data mining predictive models.
\newblock \emph{Journal of the Royal Statistical Society: Series A (Statistics
  in Society)}, 176\penalty0 (2):\penalty0 565--584, 2013.

\bibitem[Wagstaff(2012)]{Wagstaff12}
Wagstaff, K.~L.
\newblock Machine learning that matters.
\newblock In \emph{Proceedings of the International Conference on Machine
  Learning (ICML)}, pp.\  1851–1856, 2012.

\bibitem[Wang et~al.(2022{\natexlab{a}})Wang, Han, Patel, and
  Rudin]{WangHanEtAl2022}
Wang, C., Han, B., Patel, B., and Rudin, C.
\newblock {In Pursuit of Interpretable, Fair and Accurate Machine Learning for
  Criminal Recidivism Prediction}.
\newblock \emph{Journal of Quantitative Criminology}, pp.\  1--63,
  2022{\natexlab{a}}.

\bibitem[Wang et~al.(2020)Wang, Huang, Gao, Zhou, Lai, Li, Xian, Qian, Li,
  Huang, et~al.]{wang2020initial}
Wang, F., Huang, S., Gao, R., Zhou, Y., Lai, C., Li, Z., Xian, W., Qian, X.,
  Li, Z., Huang, Y., et~al.
\newblock Initial whole-genome sequencing and analysis of the host genetic
  contribution to {COVID}-19 severity and susceptibility.
\newblock \emph{Cell Discovery}, 6\penalty0 (1):\penalty0 83, 2020.

\bibitem[Wang et~al.(2017)Wang, Rudin, Doshi-Velez, Liu, Klampfl, and
  MacNeille]{WangEtAl2017}
Wang, T., Rudin, C., Doshi-Velez, F., Liu, Y., Klampfl, E., and MacNeille, P.
\newblock A {B}ayesian framework for learning rule sets for interpretable
  classification.
\newblock \emph{Journal of Machine Learning Research}, 18\penalty0
  (70):\penalty0 1--37, 2017.

\bibitem[Wang et~al.(2021)Wang, Kale, Nori, Stella, Nunnally, Chau, Vorvoreanu,
  Vaughan, and Caruana]{wang2021gam}
Wang, Z.~J., Kale, A., Nori, H., Stella, P., Nunnally, M., Chau, D.~H.,
  Vorvoreanu, M., Vaughan, J.~W., and Caruana, R.
\newblock Gam changer: Editing generalized additive models with interactive
  visualization.
\newblock \emph{Advances in Neural Information Processing Systems, Bridging the
  Gap: From Machine Learning Research to Clinical Practice (Research2Clinics)
  Workshop}, 2021.

\bibitem[Wang et~al.(2022{\natexlab{b}})Wang, Zhong, Xin, Takagi, Chen, Chau,
  Rudin, and Seltzer]{wang2022timbertrek}
Wang, Z.~J., Zhong, C., Xin, R., Takagi, T., Chen, Z., Chau, D.~H., Rudin, C.,
  and Seltzer, M.
\newblock Timbertrek: Exploring and curating sparse decision trees with
  interactive visualization.
\newblock In \emph{2022 {IEEE} Visualization and Visual Analytics (VIS)}, pp.\
  60--64. IEEE, 2022{\natexlab{b}}.

\bibitem[Watson-Daniels et~al.(2023)Watson-Daniels, Parkes, and
  Ustun]{watson2023predictive}
Watson-Daniels, J., Parkes, D.~C., and Ustun, B.
\newblock Predictive multiplicity in probabilistic classification.
\newblock In \emph{Proceedings of the AAAI Conference on Artificial
  Intelligence}, volume~37, pp.\  10306--10314, 2023.

\bibitem[Wikipedia(2024)]{RightToExpl}
Wikipedia.
\newblock Right to explanation.
\newblock \url{https://en.wikipedia.org/wiki/Right_to_explanation}, 2024.

\bibitem[Xin et~al.(2022)Xin, Zhong, Chen, Takagi, Seltzer, and
  Rudin]{xin2022exploring}
Xin, R., Zhong, C., Chen, Z., Takagi, T., Seltzer, M., and Rudin, C.
\newblock Exploring the whole {R}ashomon set of sparse decision trees.
\newblock In \emph{Neural Information Processing Systems ({NeurIPS})},
  volume~35, pp.\  14071--14084, 2022.

\bibitem[Yanagawa \& Sato(2024)Yanagawa and Sato]{YanagawaEtAl2024}
Yanagawa, M. and Sato, J.
\newblock Seeing is not always believing: Discrepancies in saliency maps.
\newblock \emph{Radiology: Artificial Intelligence}, 6\penalty0 (1):\penalty0
  e230488, 2024.

\bibitem[Zeng et~al.(2017)Zeng, Ustun, and Rudin]{ZengUsRu2017}
Zeng, J., Ustun, B., and Rudin, C.
\newblock Interpretable classification models for recidivism prediction.
\newblock \emph{Journal of the Royal Statistical Society: Series {A}
  (Statistics in Society)}, 180\penalty0 (3):\penalty0 689--722, 2017.

\bibitem[Zhong et~al.(2023)Zhong, Chen, Liu, Seltzer, and
  Rudin]{zhong2023exploring}
Zhong, C., Chen, Z., Liu, J., Seltzer, M., and Rudin, C.
\newblock Exploring and interacting with the set of good sparse generalized
  additive models.
\newblock In \emph{Neural Information Processing Systems ({NeurIPS})}, 2023.

\bibitem[Zhu et~al.(2023)Zhu, Tian, Semenova, Liu, Xu, Scarpa, and
  Rudin]{zhu2023GroupFasterRisk}
Zhu, C.~Q., Tian, M., Semenova, L., Liu, J., Xu, J., Scarpa, J., and Rudin, C.
\newblock Fast and interpretable mortality risk scores for critical care
  patients.
\newblock \emph{arXiv preprint arXiv:2311.13015}, 2023.

\end{thebibliography}
\bibliographystyle{icml2024}

\newpage
\appendix
\onecolumn

\section{Metrics to Gauge the Rashomon Effect}\label{app:metrics}
There are many different ways to assess the Rashomon Effect.
\begin{itemize}
\item For measuring the \textbf{size} of Rashomon set: TreeFARMS \citep{xin2022exploring} computes the number of sparse decision trees. The GAM Rashomon set algorithm \citep{zhong2023exploring} computes the number of unique support sets (for GAMs, including linear and additive models); the Rashomon set includes a convex set of models for each support set. CorelsEnum of \citet{mata2022computing} enumerates the Rashomon set of rule lists, while DiversiTree of \citet{ahanor2022diversitree} gives a set of diverse, close-to-optimal mixed integer programming solutions. For ridge regression, the size of the Rashomon set can be computed in closed-form \citep{SemenovaRuPa2022}.
\item For measuring \textbf{diversity of predictions} (for classifiers), we can use the pattern diversity metric of \citet{SemenovaRuPa2022} or the pairwise disagreement of \citet{black2022model}. The ambiguity and discrepancy metrics of \citet{marx2020predictive} can further help to understand the conflicting predictions from the Rashomon set's models. For example, ambiguity tells how many people's bail decision could be changed by using a different model from the Rashomon set, while discrepancy tells us the model in the Rashomon set with the most bail decisions changed relative to a baseline (deployed) model.  The Hacking Interval framework of \citet{coker2021theory} contains calculations that show maximum and minimum predictions within the Rashomon set for several different types of algorithms.
\item For measuring \textbf{variable importance diversity}, we can use Model Class Reliance of \citet{fisher2019all} or \citet{smith2020model} to get a range of variable importance values in the Rashomon set. We can visualize the ``cloud'' of variable importance using the approach of \citet{dong2020exploring}, which plots each model in variable importance space.
\item For \textbf{probabilistic classification}, the Rashomon capacity metric of \citet{hsu2022rashomon} can be used, or probabilistic ambiguity/discrepancy of \citet{watson2023predictive}.
\item The Rashomon ratio or pattern Rashomon ratio of \citet{SemenovaRuPa2022}, as well as the fraction of good models in the hypothesis space, can help to understand the \textbf{simplicity of the learning problem}.
\end{itemize}
\end{document}